\documentclass[10pt,twocolumn,letterpaper]{article}

\usepackage[pagenumbers]{cvpr}
\usepackage{graphicx}
\usepackage{amsmath}
\usepackage{amssymb}
\usepackage{booktabs}
\usepackage{xspace}
\usepackage{interval}
\usepackage{bm}
\usepackage{mathtools}
\usepackage{footmisc}
\usepackage{booktabs}
\usepackage{multirow}
\newcommand{\hdrule}{\midrule[\heavyrulewidth]}

\usepackage{siunitx}
\usepackage{bbold}
\usepackage{color}
\definecolor{citecolor}{RGB}{0, 113, 188}
\usepackage[pagebackref=true,breaklinks=true,letterpaper=true,colorlinks,citecolor=citecolor, bookmarks=false]{hyperref}

\usepackage[capitalize]{cleveref}

\crefname{section}{Sec.}{Secs.}
\Crefname{section}{Section}{Sections}
\Crefname{table}{Table}{Tables}
\crefname{table}{Tab.}{Tabs.}

\begin{document}

\title{Self-Supervised Learning for Semi-Supervised Temporal Language Grounding}

\author{Fan Luo,~
Shaoxiang Chen,~
Jingjing chen,~
Zuxuan Wu,~
Yu-Gang Jiang\\
Shanghai Key Lab of Intelligent Information Processing, School of Computer Science, Fudan University\\
}
\maketitle
\begin{abstract}
Given a text description, Temporal Language Grounding (TLG) aims to localize temporal boundaries of the segments that contain the specified semantics in an untrimmed video. TLG is inherently a challenging task, as it requires comprehensive understanding of both sentence semantics and video contents.
Previous works either tackle this task in a fully-supervised setting that requires a large amount of temporal annotations or in a weakly-supervised setting that usually cannot achieve satisfactory performance.
Since manual annotations are expensive, to cope with limited annotations, we tackle TLG in a semi-supervised way by incorporating self-supervised learning, and propose \textbf{S}elf-\textbf{S}upervised \textbf{S}emi-\textbf{S}upervised \textbf{T}emporal \textbf{L}anguage \textbf{G}rounding (S$^4$TLG).
S$^4$TLG consists of two parts:
(1) A pseudo label generation module that adaptively produces instant pseudo labels for unlabeled samples based on predictions from a teacher model;
(2) A self-supervised feature learning module with inter-modal and intra-modal contrastive losses to learn video feature representations under the constraints of video content consistency and video-text alignment.
We conduct extensive experiments on the ActivityNet-CD-OOD and Charades-CD-OOD datasets. The results demonstrate that our proposed S$^4$TLG can achieve competitive performance compared to fully-supervised state-of-the-art methods while only requiring a small portion of temporal annotations.
\end{abstract}

\vspace{-6mm}
\section{Introduction}
\label{sec:intro}
\noindent Video understanding is a challenging task in computer vision. Many related tasks have been extensively studied due to its wide applications, such as video captioning~\cite{Chen2020}, action localization~\cite{Shou2016} and video question answering~\cite{Tapaswi2016}. Temporal language grounding (TLG)~\cite{Gao2017,Hendricks2017}, which aims at automatically locating the temporal boundaries of the segments indicated by natural language descriptions in an untrimmed video, has recently attracted increasing attention from both the computer vision and the natural language processing communities because of its potential applications in video search engines and automated video editing.

The state-of-the-art models for tackling the temporal language grounding task in a fully-supervised setting can be roughly categorized into two groups:~\textit{proposal-based} models~\cite{Gao2017,Ge2019,Jiang2019,Liu2018,Liu2018a,Liu2021,Zhang2020} and~\textit{regression-based} (proposal-free) models~\cite{Yuan2019a,Zeng2020,Opazo2020,Chen2020a,Wang2019,Mun2020,Nan2021,Zhao2021,Zhou2021,Zhang2021}.
Their major difference is the way of generating temporal predictions and training/inference efficiency (\eg, memory consumption and convergence rate).
But all of these methods require precise temporal boundaries of the sentence in the video for model training.
Since manually annotating such temporal labels is expensive and time-consuming, it is extremely challenging to scale up such fully-supervised methods.

To alleviate the annotation cost, weakly-supervised and semi-supervised learning methods are common solutions. Weakly-supervised temporal language grounding methods require only video-sentence pairs for training, that is, without the requirement for temporal boundary annotations.
However, recent studies have shown that the overall performance of weakly-supervised methods~\cite{Duan2018,Gao2019,Mithun2019,Song2020,Tan2019,Zhou2021a} are greatly inferior to fully-supervised methods in temporal language grounding. 
Compared with weakly-supervised learning, semi-supervised learning can make use of a limited amount of annotated data as necessary supervision to achieve higher performance. 
Thus, semi-supervised methods could be a more appropriate solution for problems where annotations are expensive but not completely impossible to obtain. Though semi-supervised learning has been widely explored in tasks such as object detection~\cite{Tang2021,Zhou2021}, temporal action localization~\cite{Ji2019,Wang2021}, to the best of our knowledge, we are the first to explore semi-supervised learning for temporal language grounding.

The major challenge in semi-supervised temporal language grounding is how to learn from the large amount of coarsely annotated data.
One common practice for semi-supervised learning is generating pseudo labels, in TLG, we have to handle different forms of labels as previously mentioned, e.g. proposal scores or boundary coordinates.
Besides generating pseudo labels, learning a proper feature representation is also crucial for this task and is also our focus.
Recent progress of self-supervised learning suggests that preserving consistency between different views of samples (generated by perturbation/augmentation) is an effective way to make use of unlabeled data~\cite{Miyato2019,li2021motion,behrmann2021long,wang2021removing,patrick2021space
}.
However, these methods mostly focus on the intra-modal consistency alone and can not make full use of the multi-modal (video and text) data in TLG, which could lead to sub-optimal performance. 
For the TLG task, it is essential to take into consideration the rich textual information and perform inter-modal feature learning.
Moreover, it is computationally expensive to perform perturbation/augmentation in the pixel space, thus we need to design proper perturbations in the feature space for this task to facilitate efficient contrastive feature learning.

The goal of this paper is to tackle the above challenges.
Our proposed framework correspondingly consists of two parts: pseudo label generation and self-supervised feature learning.
The pseudo label generation aims at producing more precise pseudo labels for different types of grounding models. In the semi-supervised training process, we adopt the teacher-student scheme to use a teacher model to produce instant pseudo labels for the student model to leverage unlabeled data.
The self-supervised feature learning includes intra-modal and inter-modal contrastive learning, for learning video feature representations under the constraint of video content consistency and video-text alignment, respectively. 
To produce different views of a sample, we design two types of sequential data perturbations: time lagging and video compression simulation, which can make the learned video features more robust and also focus more on the semantics of video content to preserve consistency within the video modality and between the video and text modalities.
Our proposed framework is abbreviated as S$^4$TLG (\textbf{S}elf-\textbf{S}upervised \textbf{S}emi-\textbf{S}upervised \textbf{T}emporal \textbf{L}anguage \textbf{G}rounding).

In summary, our main contributions are summarized as follows:
1) To the best of our knowledge, we are the first to study semi-supervised temporal language grounding and propose a general framework that works well for both regression-based and proposal-based grounding models.
2) We have specifically designed inter-modal and intra-modal contrastive losses for self-supervised feature learning in temporal language grounding, which can effectively improve grounding performance.
3) We extensively tested the proposed S$^4$TLG on two challenging out-of-distribution datasets and achieve results comparable to fully-supervised baselines for regression-based and proposal-based models.

%------------------------------------------------------------------------
\vspace{-3mm}
\section{Related Work}
\vspace{-2mm}
\label{sec:formatting}

\textbf{Temporal language grounding} aims to locate the temporal boundary of events in untrimmed videos given language queries ~\cite{Gao2017,Hendricks2017}. Existing methods for temporal language grounding can be mainly categorized into two groups.
Proposal-based methods~\cite{Gao2017,Hendricks2017,Liu2018,Liu2018a,Ge2019,Jiang2019,Liu2021,Wang2020,Zhang2020} retrieve the most confident proposal from a set of candidates.
Initially, \cite{Gao2017} follows a two-stage paradigm, which first generates candidates by temporal sliding windows and then matches language queries with these candidates. \cite{Zhang2020} designs a fine-grained and flexible 2D temporal feature map proposal generation mechanism to capture relations among temporal segments.
Although those methods achieve promising performance, they are limited by the need of extra computational cost and high-quality ground-truth labels.
Regression-based methods~\cite{Yuan2019a,Zeng2020,Opazo2020,Chen2020a,Wang2019,Mun2020,Nan2021,Zhao2021} are proposed to solve temporal language grounding task in an end-to-end manner.
Some methods~\cite{Yuan2019a} focus on cross-modal attention, while others propose to learn fine-grained cross-modal interactions with a segment tree~\cite{Zhao2021} or local-global context~\cite{Mun2020}. \cite{Zhang2021} attempts to address grounding in a span-based question answering way.
Since annotations are expensive, significant effort has been devoted to solving temporal language grounding problem in a weakly-supervised way recently~\cite{Duan2018,Gao2019,Mithun2019,Zhang2020a,Zhang2020b,Ma2020}.
The mainstream approaches mostly follow the multiple instance learning paradigm~\cite{Mithun2019}. Apart from this, \cite{Zhang2020a} proposes to solve the weakly-supervised temporal language grounding based on contrastive learning.

\textbf{Semi-supervised learning (SSL)} is a dominant approach in machine learning to facilitate learning with limited labeled data and large amounts of unlabeled data.
SSL has been extensively studied~\cite{Arazo2020,Xie2020,Sohn2020,Pham2021,Laine2017,Tarvainen2017} in image classification. Their focus can be roughly categorized into two groups: pseudo label generation and consistency regularization. Pseudo label generation produces artificial labels with a teacher model for unlabeled data and using them to further train the student model. \cite{Arazo2020} proposes to train the student model with a fixed pre-trained teacher model. Subsequent studies investigate the usage of pseudo labels. For instance, \cite{Xie2020} proposes Noisy-Student to train a student model with an iterative teacher model; \cite{Sohn2020} proposes FixMatch to apply different levels of augmentation to pseudo label generation and prediction; \cite{Pham2021} updates the teacher model from the feedback of the student model.
Consistency regularization methods encourage models to generate consistent outputs from perturbed input data and have been widely used in SSL.
\cite{Laine2017} maintains a regularization for ensemble prediction.
\cite{Tarvainen2017} updates the teacher model by averaging weights of the student model at each training step so as to promote the quality of labels.
Most related to our work are two previous methods that study semi-supervised temporal action localization (SSTAL).
\cite{Ji2019} designs two sequential perturbations based on the mean teacher framework,
\cite{Wang2021} proposes to incorporate self-supervised learning into SSTAL. But language grounding is much more complicated than action localization and requires inter-modal joint learning.

\begin{figure*}
	\centering
	\includegraphics[width=.95\textwidth]{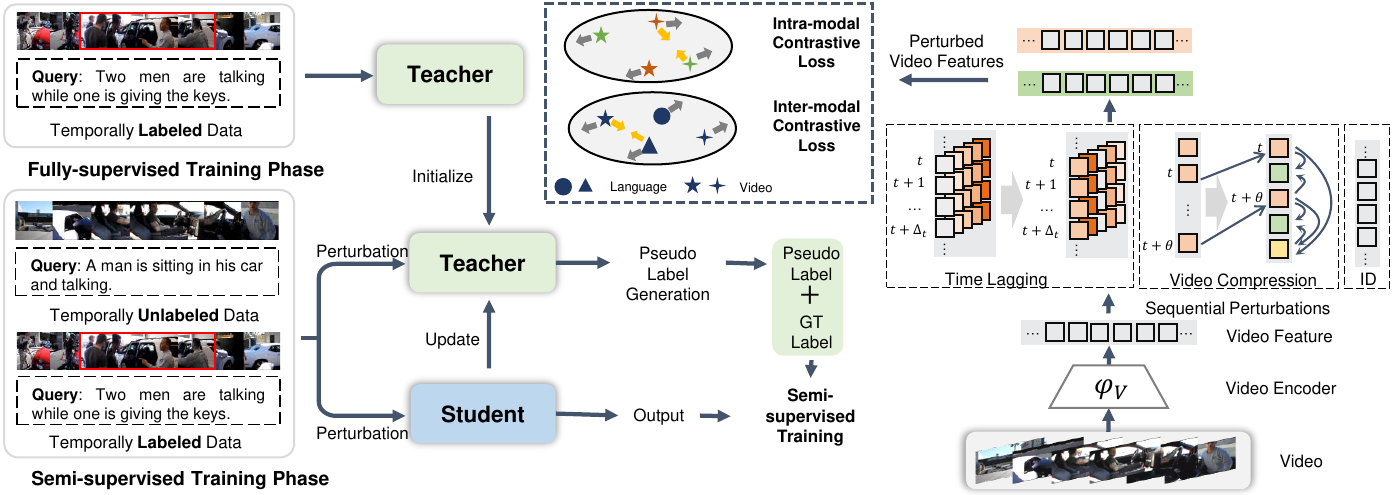}
	\caption{Overview of our S$^4$TLG. In the fully-supervised training phase, we train the initial teacher model with limited labeled data. Then in the semi-supervised training phase, the teacher model generates pseudo labels for the unlabeled data to train the student model, and gets instant update from the student model in each epoch. Sequential perturbation is applied to semi-supervised learning as data augmentation. Meanwhile the contrastive feature learning also relies on the perturbations to generate different views. The intra-modal contrastive loss forces the model to focus more on video semantics, and the inter-modal contrastive loss tries to align the video and sentence in the feature space to guide the aggregated video feature towards having more correct semantics.}
	\label{fig:framework}
	\vspace{-4mm}
\end{figure*}

\textbf{Self-supervised learning} leverages unlabeled data to make the model learn intrinsic knowledge from data.
Previous methods rely on pretext tasks to learn better feature representations---representative ones are order verification~\cite{Misra2016}, sequence sorting~\cite{Lee2017}, masked feature reconstruction and clip-order prediction~\cite{Wang2021}.
Recently, contrastive learning methods~\cite{Pathak2016,Han2019,chen2020simple,chen2020big,he2019moco,chen2020mocov2,li2021motion,behrmann2021long,wang2021removing,patrick2021space}, have attracted increasing attention since they are proven to be able to learn more powerful feature representations for downstream tasks. Its basic idea is to align similar samples in feature space and push away the dissimilar samples, so that the learned features can be discriminative.

%-------------------------------------------------------------------------
\vspace{-4mm}
\section{Our Approach}
\vspace{-2mm}

In this section, we first give the problem definition of the semi-supervised temporal language grounding task.
Then we introduce the details of our proposed S$^4$TLG framework, which includes pseudo label generation, sequential perturbation, and self-supervised feature learning.
\vspace{-2mm}
\subsection{Problem Formulation}
\vspace{-2mm}
Given an untrimmed video $\mathbf{V} \in \mathbb{R}^{T}$ and a corresponding sentence description $\mathbf{S} \in \mathbb{R}^{N}$, where $T$ is the length of video sequence and $N$ is the number of words,
temporal language grounding aims at locating the video segment $\tau=[s,e]$ which is mostly related to the sentence. In the semi-supervised setting, only a part of the video-sentence pairs are provided with temporal boundary ground truth while the remaining ones are not available during training.
The grounding model can be formulated as:
\begin{equation}
	\tau = f(\mathbf{V},\mathbf{S}),
\end{equation}
where $f(\cdot,\cdot)$ is usually a composite of feature encoding, cross-modal processing, segment predicting, and post-processing steps. Since the focus of this paper is the learning process, so we use \textit{vanilla} grounding models instead of advanced ones.
%-------------------------------------------------------------------------
\subsection{Pseudo Label Generation}

The loss functions of temporal language grounding can take different forms, requiring different methods for generating pseudo labels. Below, we discuss how to derive pseudo labels for regression-based grounding and proposal-based grounding, respectively.

\textbf{Regression-based Model.}
Generally, regression-based grounding models can be formulated as:
\begin{equation}
	\begin{split}
		\tau_{r} &= f_{r}(\mathbf{V},\mathbf{S}),\\
		f_{r}(\mathbf{V},\mathbf{S}) &= \mathrm{\Omega}_{r}(\mathrm{\Phi}_{int}(\varphi_{V}(\mathbf{V}),\varphi_{S}(\mathbf{S}))),
	\end{split}
\end{equation}
where $\mathrm{\Omega}_{r}(\cdot)$ denotes the temporal boundary regression module, $\mathrm{\Phi}_{int}(\cdot,\cdot)$ is the cross-modal interaction module, and  $\varphi_{V}(\cdot)$ and $\varphi_{S}(\cdot)$ are the feature encoding modules for the video and sentence, respectively.
Following previous work~\cite{Yuan2019a}, $\varphi_{V}$ is comprised of a CNN and a bi-directional LSTM unit, $\varphi_{S}$ is comprised of a Glove embedding layers and a bi-directional LSTM unit.
$\mathrm{\Phi}_{int}$ denotes an iterative cross-modal co-attention operation:
\begin{equation}
	\bm{h}_{r}, \bm{a} = \mathrm{\Phi}_{int}(\varphi_{V}(\mathbf{V}),\varphi_{S}(\mathbf{S})),
\end{equation}
where $\bm{h}_{r} \in \mathbb{R}^{T\times D_{h}}$ is the attended video feature and $\bm{a} \in \mathbb{R}^{T}$ is the corresponding attention weights.
Then the temporal segment is predicted as:
\begin{equation}
	\tau_{r} = [s,e] = \mathrm{\Omega}_{r}(\bm{h}_{r}),
\end{equation}
where $\mathrm{\Omega}_{r}(\cdot)$ can be a multi-layer perceptron.

During training, there are two types of loss functions:
\begin{equation}
	\mathcal{L}^{task}_{r} = \mathcal{L}^{reg}_{r}+\alpha_{r} \mathcal{L}^{cal}_{r}.
\end{equation}
$\mathcal{L}^{reg}_{r}$ is the temporal boundary regression loss, which is the smooth L1 loss
\begin{equation}
	\mathcal{L}^{reg}_{r}(\tau_{r}, \hat{\tau}_{r}) = \mathtt{SmoothL1}(s,\hat{s}) + \mathtt{SmoothL1}(e,\hat{e}),
\end{equation}
where $\hat{\tau}_{r}$ is the ground truth segment label\footnote{We denote ground truth labels by symbols with a hat (e.g., $\hat{\tau}_{r}$), and pseudo labels by symbols with an asterisk (e.g., $\tau_{r}^*$).}.
$\mathcal{L}^{cal}_{r}$ is the attention calibration loss to encourage the alignment between the attention weights and the ground truth $\hat{\bm{a}} \in \mathbb{R}^{T}$:
\begin{equation}
	\mathcal{L}^{cal}_{r}(\hat{\bm{a}}, \bm{a}) = - \frac{\sum_{t=1}^{T}{\hat{\bm{a}}_{t}\log{\bm{a}_{t}}}}{\sum_{i=t}^{T}{\hat{\bm{a}}_{t}}},
\end{equation}

where $\hat{\bm{a}}_{t}$ is 1 if $s \le t \le e$, and 0 otherwise.

We design pseudo labels for both types of losses. $\tau_{r}^*$ and $\bm{a}^*$ correspond to ground truth labels $\hat{\tau}_{r}$ and $\hat{\bm{a}}$.
We directly use the teacher-model's boundary prediction $\tilde{\tau}_{r}$ as the pseudo label $\tau^{*}_{r}$, and then construct attention weight pseudo label $\bm{a}^{*}$ based on $\tau^{*}_{r}$.
At time step $t$, the attention weight is expected to be close to 1 if $s^* \le t \le e^*$, and otherwise it should be close to 0:
\begin{equation}
	\begin{split}
		\tau^*_{r}  & = [s^*,e^*] = \tilde{\tau}, \\
		\bm{a}_{t}^* & = \begin{cases}
			1 & s^* \le t \le e^*, \\
			0 & otherwise.
		\end{cases}
	\end{split}
\end{equation}

\textbf{Proposal-based Model.}
The proposal-based methods can be formulated as bellow:
\begin{equation}
	\begin{split}
		\omega_{p} &= f_{p}(\mathbf{V},\mathbf{S}),\\
		f_{p}(\mathbf{V},\mathbf{S}) &= \mathrm{\Omega}_{p}(\mathrm{\Phi}_{int}(\varphi_{V}(\mathbf{V}),\varphi_{S}(\mathbf{S}))).
	\end{split}
\end{equation}
In contrast to regression-based methods, the output of $f_p(\cdot,\cdot)$ are the candidate proposals, i.e., $\omega_{p} \in \mathbb{R}^{T\times K}$ represents $K$ anchors at each of the $T$ time steps.
And the final segment $\tau_{p}$ is selected for $\omega_{p}$ with a non-maximum suppression (NMS) process.
To make this a fair comparison, the feature encoding module $\varphi_{V}$ and $\varphi_{S}$ are kept consistent with the above regression-based method.
The cross-modal interaction module $\mathrm{\Phi}_{int}$ in the proposal-based model firstly fuses multi-modal feature with a linear transformation, then applies self-attention to generate the contextual integration feature $\mathbf{h}_{p}\in \mathbb{R}^{T\times D_{h}}$:
\begin{equation}
	\mathbf{h}_{p} = \mathrm{\Phi}_{int}(\mathbf{V},\mathbf{S}).
\vspace{-1mm}
\end{equation}
Following \cite{Wang2020}, the proposal generation module $\mathrm{\Omega}_{p}$ is a combination of an anchor prediction submodule and a boundary prediction submodule:
\begin{equation}
	\mathbf{c}, \mathbf{b} = \mathrm{\Omega}_{p}(\mathbf{h}_{p}),
\end{equation}
where $\mathbf{c} \in [0,1]^{T \times K}$ is the confidence score of each candidate proposal, and $\mathbf{b} \in [0,1]^{T}$ represents the probability of being a boundary point. During inference, the anchor score and boundary probability are jointly used to produce the final score of a candidate proposal.
The proposal-based loss $\mathcal{L}^{task}_{p}$ is then defined as below:
\begin{equation}
	\mathcal{L}^{task}_{p} = \mathcal{L}^{anchor}_{p} + \alpha_{p}\mathcal{L}^{boundary}_{p}.
\end{equation}
The anchor loss $\mathcal{L}^{anchor}_{p}$ is calculated with the prediction $\bm{c}$ and label: $\hat{\mathbf{c}} \in \mathbb{R}^{T \times K}$
\begin{align}
	\mathcal{L}^{anchor}_{p}(\hat{\mathbf{c}},\mathbf{c}) = - & \sum_{t=1}^{T}\sum_{k=1}^{K} w_{k}^{+}\hat{\bm{c}}_{tk}\log{\bm{c}_{tk}}+ \\ \nonumber
	                                                          & w_{k}^{-}(1-\hat{\bm{c}}_{tk})\log{(1-\bm{c}_{tk})},
\end{align}
where $\hat{c}_{tk}$ is 1 if the corresponding proposal has a temporal IoU with the ground truth segment $\hat{\tau}$ larger than 0.5 and 0 otherwise,
and the $w_{k}^{+}, w_{k}^{-}$ are the positive and negative sample weights of the $k$-th proposal.
The boundary loss $\mathcal{L}^{boundary}_{p}$ is calculated with the label $\hat{\mathbf{b}} \in \mathbb{R}^{T}$, where $\hat{\mathbf{b}}_{t}$ is 1 if $t$ equals $s$ or $e$, otherwise $b_{t}$ is 0. $w_b^{+}$ and $w_{b}^{-}$ are positive and negative weights. The loss is formulated as:
\begin{align}
	\mathcal{L}^{boundary}_{p}(\mathbf{b}, \hat{\mathbf{b}}) = - & \sum_{t=1}^{T} w_{b}^{+} \hat{\mathbf{b}}_{t} \log{\mathbf{b}_{t}} + \\ \nonumber
	                                                             & w_{b}^{-} (1-\hat{\mathbf{b}}_{t}) \log{(1-\mathbf{b}_{t})}.
\end{align}

Compared with the regression-based method, there are a set of candidate proposals $\omega_{p}$ in the proposal-based model. We firstly select the candidate with the highest score as the pseudo label $\tau^{*}_p$, then the boundary pseudo label $\mathbf{b}^{*}$ is defined as:
\begin{equation}
	\mathbf{b}_{t}^{*} = \begin{cases}
		1 & t = s^{*} \ \text{or} \ t = e^{*}, \\
		0 & \text{otherwise}.
	\end{cases}
\end{equation}
As for the anchor pseudo label $\mathbf{c}^{*}$, we first compute the IoU scores $\text{score}_{tk}$ of each anchor with the pseudo segment label $\tau^{*}_p$, and then the anchor pseudo labels are assigned as
\begin{equation}
	\mathbf{c}_{tk}^{*} = \begin{cases}
		1 & \text{score}_{tk} \ge 0.5, \\
		0 & \text{score}_{tk} < 0.5.
	\end{cases}
\end{equation}

%-------------------------------------------------------------------------
\subsection{Sequential Perturbation}
In the literature, applying stochastic perturbations to the inputs has been proven crucial for semi-supervised models~\cite{Ji2019,Wang2021} and is also widely adopted by (self-supervised) contrastive learning methods~\cite{Pathak2016,Han2019}. We also consider applying sequential perturbations for both semi-supervised learning and self-supervised learning.

In contrast to simply adding gaussian noise to feature maps in Mean Teacher~\cite{Tarvainen2017}, we want to further utilize the inherent property of sequence data (i.e., video).
Intuitively, due to the redundancy and temporal semantic consistency of video, moderate sequential perturbations such as short-term lagging and lossy compression would not change the semantics of a video.
For example, people can hardly feel the difference between the original video and the video with lossy compression or occasional lagging, and can still obtain the same semantic information.
Based on this insight, for the sequential video feature $\bm{V}\in \mathbb{R}^{T \times D}$, we design two essential types of sequential perturbations\footnote{The perturbations are applied to the video features instead of the video frames, since the latter makes the computational cost much higher.}: time lagging and video compression simulation.

\textbf{Time lagging.} As shown in Figure~\ref{fig:framework}, the time lagging perturbation performs random temporal disturbance to a feature sequence.
Given a sequential video feature $\bm{V}$, we randomly select a small segment $r=[r_s,r_e], (1\le r_s \le r_e \le T, \frac{(r_e-r_s)}{T}\le 4\%)$ and randomly shuffle the features along the channel dimension within the segment. We keep the channel permutation temporally consistent to avoid overly perturbing the semantics.
Previous work~\cite{Wang2021} performs temporal feature flipping which flips the whole feature sequence along the temporal dimension. The difference between temporal feature flipping and time lagging is that temporal feature flipping reverses the whole feature while the time lagging randomly selects a small local segment and shuffles those features along the channel dimension. Thus time lagging retains more semantic information and temporal structure than temporal feature flipping. ~\cite{Wang2021} is applied to temporal localization of actions, which is usually less complex and less sensitive to temporal flipping.
In the temporal language grounding task, however, the temporal structure is critical for localizing the sentence and flipping the whole video sequence can dramatically change the semantics of a video.
Thus our designed time lagging is a more suitable perturbation for TLG.

\textbf{Video compression simulation.} We also simulate lossy video compression in the feature space as a type of perturbation since it generally preserves the essential contents. 
Inspired by the group of pictures structure (I, B, P frames) in video compression standards~\cite{watkinson2012mpeg}, we define three types of features that are constructed differently.
Given a target compression rate $\theta$, I-feature is obtained from the original video feature by uniform sampling with step size $\frac{1}{\theta}$, and is used as the reference feature afterwards.
P-feature is obtained by mean pooling the two preceding reference features,
and B-feature is obtained by the mean pooling of the nearest two reference features before and after it.
The reference features for P-feature is I-feature, and the referenced features for B-feature can be I-feature or P-feature.
The resulting features can be regarded as video event representation with different compression rates but the same semantics, and increasing the compression rate leads to more attention being paid to the overall video semantics.
%-------------------------------------------------------------------------

As shown in Figure~\ref{fig:framework}, during semi-supervised learning, the sequential perturbation is used to augment the input data for both teacher and student models. Meanwhile, in self-supervised learning (Sec~\ref{sec:ssfl}), sequential perturbations are also used to generate different views for given input videos, forcing the model to learn the underlying semantic features and be invariant to these perturbations.

\subsection{Self-Supervised Feature Learning}\label{sec:ssfl}

We perform self-supervised feature learning on the encoded visual and textual features.
Based on the previous two types of sequential perturbations (data augmentation), we design inter-modal and intra-modal contrastive losses to assist the S$^4$TLG to learn discriminative multi-modal feature representation.

We first give a general formulation of contrastive loss:
\begin{small}
	\begin{equation}
		\begin{split}
			&\mathcal{L}_{CL}(\mathbf{X},\mathbf{Y},\mathcal{X}^{-}, \mathcal{Y}^{-}) = \\
			& \sum_{\mathbf{X}^{-} \in \mathcal{X}^{-}}{\max{[0,\Delta - l(\mathbf{X},\mathbf{Y})+l(\mathbf{X}^{-},\mathbf{Y})]}}+\\
			& \sum_{\mathbf{Y}^{-} \in \mathcal{Y}^{-}}{\max{[0,\Delta - l(\mathbf{X},\mathbf{Y})+l(\mathbf{X},\mathbf{Y}^{-}]}},
		\end{split}
	\end{equation}
\end{small}
where $l(\cdot,\cdot)$ is a similarity function (e.g, cosine similarity),
$\mathbf{X},\mathbf{Y} \in \mathbb{R}^{D_{h}}$ is a pair of positive samples,
and $\mathcal{X}^{-}$ and $\mathcal{Y}^{-}$ are the sets of chosen negative sample for $\mathbf{X}$ and $\mathbf{Y}$, respectively.
$\Delta$ is the margin.
The goal of $\mathcal{L}_{CL}$ is to pull positive pairs closer and push negative pairs further in the embedding space.

\textbf{Inter-modal contrastive loss.} Consider a mini batch $\{\bm{V}_{i},\bm{S}_{i}\}_{i=1}^{B}$ with $B$ video and sentence features, the inter-modal contrastive loss is:
\begin{align}
	\mathcal{L}_{CL}^{inter} = \sum_{i=1}^{B}{\mathcal{L}_{CL}(\mathbf{V}_{i},\mathbf{S}_{i}, \mathcal{V}_{i}^{-},\mathcal{S}_{i}^{-})},
\end{align}
where $\mathcal{V}_{i}^{-} = \{\mathbf{V}_{j} | j\in [1,B]\ \text{and} \ j\neq i\}$ and $\mathcal{S}_{i}^{-} = \{\mathbf{S}_{j} | j\in [1,B]\ \text{and} \ j\neq i\}$ are the negative samples for the $i$-th video and sentence pair.
The inter-modal contrastive loss encourages the semantically related video and sentence to be similar, and decreases the similarity between semantically irrelevant video sentence pairs.

\textbf{Intra-modal contrastive loss.}
To encourage the model to focus more on the semantics of the video, we also apply contrastive learning on augmented versions of video features.
\begin{equation}
	\begin{split}
		\mathbf{A}_{i1}, \mathbf{A}_{i2} &= \mathtt{Aug}(\mathbf{V}_i), \\
		\mathcal{L}_{CL}^{intra} &= \sum_{i=1}^{B}{\mathcal{L}_{CL}(\mathbf{A}_{i1},\mathbf{A}_{i2},\mathcal{A}_{i1}^{-},\mathcal{A}_{i2}^{-})},
	\end{split}
\end{equation}
where $\mathtt{Aug}(\cdot)$ takes a video sample and augments it twice using perturbations chosen randomly from the set \{time lagging, video compression simulation, identity mapping\}, and $\mathbf{A}_{i1}, \mathbf{A}_{i2} \in \mathbb{R}^{T\times D_{h}}$ are the augmented video features.
Similar to inter-modal contrastive loss, intra-modal contrastive loss also regards the rest of $B-1$ samples as negative, that $\mathcal{A}_{i1}^{-} = \{\mathbf{A}_{j1}|j \in [1,B] \ \text{and} \ j\neq i\}, \mathcal{A}_{i2}^{-} = \{\mathbf{A}_{j2}|j \in [1,B] \ \text{and} \ j\neq i\}$ are the negative samples for sample $i$.

The overall self-supervised loss is:
\begin{equation}
	\mathcal{L}^{self} = \mathcal{L}_{CL}^{inter} + \mathcal{L}_{CL}^{intra}.
\end{equation}

\subsection{The Training Process}
There are two phases in our grounding model during the training process.
In the first phase, we pre-train the model with a limited amount of annotated data in a supervised manner to obtain a teacher model.
In the second phase, the teacher model generates pseudo labels for the unlabeled data in each epoch.
Following the instant-teaching~\cite{Zhou2021} framework, we use instant pseudo labeling in our S$^4$TLG.
We then perform the semi-supervised training by optimizing the semi-supervised loss $\mathcal{L}_{task}$ and self-supervised loss $\mathcal{L}_{self}$ using all the training samples:
\begin{equation}
	\mathcal{L}^{all} = \mathcal{L}^{task} + \beta \mathcal{L}^{self},
\end{equation}
where $\mathcal{L}^{task}$ can be $\mathcal{L}^{task}_r$ or $\mathcal{L}^{task}_p$, depending on the type of grounding model being trained, and $\beta$ is a hyper parameter that controls the weight of self-supervised loss.

%-------------------------------------------------------------------------
\vspace{-3mm}
\section{Experiments}
\vspace{-2mm}
%-------------------------------------------------------------------------
\subsection{Datasets}
Recent studies~\cite{Yuan2021,otani2020challengesmr} have suggested that the data distribution of popular datasets used by current TLG methods are mostly biased and can not reflect the actual grounding capabilities of state-of-the-art models. This problem can have even greater negative effects in our semi-supervised setting. Thus we decide to follow the settings in~\cite{Yuan2021} and use a novel split of existing datasets, in which the training and testing data are designed to have different distributions, so that the model must learn intrinsic clues from the video and sentence.

\textbf{Charades-CD-OOD.} Charades-CD-OOD is built on the Charades-STA dataset~\cite{Gao2017}, which annotates sentence descriptions in a semi-automatic way. It contains 9,848 videos and 16,128 sentence-moment pairs. On average, the video and sentence length on the Charades-CD dataset are 29.8 seconds and 8.6 words respectively. \cite{Yuan2021} propose to split sentence-moment pairs into training, validation, and test-ood set of sizes 11,071, 859 and 3,375, respectively.

\textbf{ActivityNet-CD-OOD.} ActivityNet-CD-OOD is built on the ActivityNet Captions~\cite{Heilbron2015} dataset, which is originally for dense video captioning. The dataset has 19,970 videos and 17,031 sentence-moment pairs. Compared with Charades-STA, its variation of temporal segment lengths is much larger, ranging from a few seconds to several minutes. The average sentence length is 13.2 words, and the average video length is 117.74 seconds. According to~\cite{Yuan2021}, ActivityNet-CD-OOD contains 51,415, 3,321, 13,578 sentence-moment pairs for training, validation, and test-ood respectively.

The rate $\psi$, which represents the percentage of labeled data in training data, is set to 5\% in ActivityNet-CD-OOD and 30\% in Charades-CD-OOD, to ensure that these two datasets can have similar amounts of labeled instances\footnote{Due to the space limit, we put the results of different amounts of labels and discussion about the limitations in the Supplementary Materials.}.

\begin{table*}[t]
	\centering
	\footnotesize
	\begin{tabular}{ccrrrrrr}
		\toprule
		\multicolumn{2}{c}{\multirow{2}{*}{Methods}}          & \multicolumn{3}{c}{ActivityNet-CD-OOD} & \multicolumn{3}{c}{Charades-CD-OOD}                                        \\
		\multicolumn{2}{c}{}                                  & \begin{tabular}[c]{@{}c@{}}dR@1,\\ IoU@0.3\end{tabular}         & \begin{tabular}[c]{@{}c@{}}dR@1,\\ IoU@0.5\end{tabular}      & \begin{tabular}[c]{@{}c@{}}dR@1,\\ IoU@0.7\end{tabular} & \begin{tabular}[c]{@{}c@{}}dR@1,\\ IoU@0.3\end{tabular} & \begin{tabular}[c]{@{}c@{}}dR@1,\\ IoU@0.5\end{tabular} & \begin{tabular}[c]{@{}c@{}}dR@1,\\ IoU@0.7\end{tabular}         \\
% 		\cmidrule(lr){1-2} \cmidrule(lr){3-3}
        \midrule
		\multicolumn{2}{c}{PredictAll~\cite{Yuan2021}}        & 9.01                                   & 0.00                                & 0.00                           & 27.13                          & 0.06                           & 0.00                                   \\
		\multicolumn{2}{c}{Bias-based~\cite{Yuan2021}}        & 9.21                                   & 0.26                                & 0.11                           & 9.30                           & 5.04                           & 2.21                                   \\
		\midrule
		\multicolumn{2}{c}{ACRN~\cite{Liu2018}}               & 16.06                                  & 7.58                                & 2.48                           & 44.69                          & 30.03                          & 11.89                                  \\
		\multicolumn{2}{c}{TSP-PRL~\cite{Wu2020}}             & 29.61                                  & 16.63                               & 7.43                           & 31.93                          & 19.37                          & 6.20                                   \\
		\multicolumn{2}{c}{ABLR~\cite{Yuan2019a}}             & 33.45                                  & 20.88                               & 10.03                          & 44.62                          & 31.57                          & 11.38                                  \\
		\multicolumn{2}{c}{DRN~\cite{Zeng2020}}               & 36.86                                  & 25.15                               & 14.33                          & 40.45                          & 30.43                          & 15.91                                  \\
		\midrule
		\multicolumn{2}{c}{CTRL~\cite{Gao2017}}               & 15.68                                  & 7.89                                & 2.53                           & 44.97                          & 30.73                          & 11.97                                  \\
		\multicolumn{2}{c}{SCDM~\cite{Yuan2019}}              & 31.56                                  & 19.14                               & 9.31                           & 52.38                          & 41.60                          & 22.22                                  \\
		\multicolumn{2}{c}{2D-TAN~\cite{Zhang2020}}           & 30.86                                  & 18.38                               & 9.11                           & 43.45                          & 30.77                          & 11.75                                  \\
		\midrule
		\multicolumn{2}{c}{WSSL~\cite{Duan2018}}              & 17.00                                  & 7.17                                & 1.82                           & 35.86                          & 23.67                          & 8.27                                   \\
		\midrule
		\multicolumn{1}{c}{\multirow{3}{*}{regression-based}} & baseline@$\psi$                        & 27.56                               & 15.41                          & 5.62                           & 40.99                          & 21.02                          & 6.26  \\ 
		\multicolumn{1}{c}{}                                  & S$^4$TLG@$\psi$                            & 32.40                               & 18.70                          & 8.16                           & 43.48                          & 28.26                          & 9.15  \\ 
		\multicolumn{1}{c}{}                                  & baseline@100\%                         & 35.94                               & 20.52                          & 8.57                           & 44.63                          & 28.59                          & 9.48  \\
		\midrule
		\multicolumn{1}{c}{\multirow{3}{*}{proposal-based}}   & baseline@$\psi$                        & 28.91                               & 15.44                          & 6.16                           & 38.62                          & 21.99                          & 10.54 \\ 
		\multicolumn{1}{c}{}                                  & S$^4$TLG@$\psi$                            & 31.71                               & 18.00                          & 8.08                           & 40.72                          & 24.07                          & 12.11 \\ 
		\multicolumn{1}{c}{}                                  & baseline@100\%                         & 34.09                               & 19.02                          & 8.03                           & 40.54                          & 23.89                          & 11.21 \\ \bottomrule
	\end{tabular}
	\caption{Comparisons with state-of-the-art models on ActivityNet-CD-OOD and Charades-CD-OOD using C3D features, where $\psi$ is the percentage of labeled data, is set to 5\% for ActivityNet-CD-OOD and 30\% for Charades-CD-OOD.}
	\label{tab:label1}
	\vspace{-6mm}
\end{table*}

%-------------------------------------------------------------------------
\vspace{-2mm}
\subsection{Metrics}
\vspace{-1mm}
Following the previous setting~\cite{Yuan2021}, we use the improved ``dR@n, IoU@m" as the evaluation metric:
\begin{equation*}
	\mathrm{dR}@n,\mathrm{IoU}@m = \mathrm{R}@n,\mathrm{IoU}@m \cdot \gamma^s \cdot \gamma^e,
\end{equation*}
where the ``R@n, IoU@m" is the percentage of the query sentences that our model can produce at least one appropriate moment whose IoU with the ground truth is larger than $m$ in the top-$n$ results, and $\gamma^s = 1-\left\| s - \hat{s} \right\|$, $\gamma^e = 1-\left\| e - \hat{e} \right\|$ are coefficients used to discount the boundary offsets. Both $\tau$ and $\hat{\tau}$ are normalized to $[0,1]$ by dividing the whole video length.
Specifically, we report the results for $n \in \{1,5\}$ and $m \in \{0.3,0.5,0.7\}$ on the Charades-CD-OOD and ActivityNet-CD-OOD datasets.

%-------------------------------------------------------------------------
\vspace{-2mm}
\subsection{Implementation Details}
\vspace{-1mm}
We extract the C3D~\cite{Tran2015} visual features on both datasets. The description sentences are encoded by 300d GloVe~\cite{Pennington2014} word embedding vectors. The video length $T$ is set as 128 for computing convenience, thus the video feature sequence will be uniformly sampled if length is greater than 128, otherwise zero-padded. The batch size is set to 32, in which the ratio of labeled data is $\psi$.
Adam~\cite{Kingma2015} is used as the optimizer, with a learning rate 0.001 for both regression-based method and proposal-based method. The dimension of all hidden layers is 512. Weights $\alpha_{r}$ and $\alpha_{p}$ are 0.01 and 1.0 respectively, and weight $\beta$ is 1.0. The simulated compression rate $\theta$ is randomly chosen from $\{0.25, 0.5, 0.75\}$ and margin $\Delta$ is 1.0. We take the mean of multiple experiments with different random seeds as the results.

\vspace{-2mm}
\subsection{Performance Comparison}
\vspace{-1mm}
We first compare our S$^4$TLG with state-of-the-art temporal language grounding methods on both ActivityNet-CD-OOD and Charades-CD-OOD datasets.
These SOTA methods can be categorized into several groups:
(1) Regression-based methods: \textbf{ACRN}~\cite{Liu2018}: a memory-based attentive cross-modal retrieval network; \textbf{ABLR}~\cite{Yuan2019a}: an iterative attention-based location regression network; \textbf{TSP-PRL}~\cite{Wu2020}: a coarse-to-fine tree-structured policy based progressive reinforcement learning model; \textbf{DRN}~\cite{Zeng2020}: a dense regression network with multiple regression loss.
(2) Proposal-based methods: \textbf{CTRL}~\cite{Gao2017}: Cross-modal semantics alignment temporal localizer, \textbf{SCDM}~\cite{Yuan2019}: Language guided semantic conditioned dynamic modulation network, \textbf{2D-TAN}~\cite{Zhang2020}: 2D temporal adjacent semantic relation encoded network.
(3) Weakly-supervised method: \textbf{WSSL}~\cite{Duan2018}: a weakly-supervised crossing attention multi-model localization network.
Note that the performance of these methods are reported by~\cite{Yuan2021}, these methods will mostly experience a significant performance drop on the out-of-distribution test set compared to the original test set.

\vspace{-1mm}
\textbf{Comparisons with semi-supervised baseline}.
To demonstrate the effectiveness of our proposed pseudo label generation and self-supervised feature learning, we first compare our full method with a baseline method\footnote{More details in the Supplementary Materials.}, which is obtained by removing the label generation and self-supervised feature learning from our method and is only trained with limited labeled data (baseline@$\psi$). 
To demonstrate that the proposed pseudo label generation and self-supervised feature learning can be broadly applied to different types of models, we integrate them into both regression-based and proposal-based models.
Table~\ref{tab:label1} summarizes the performance. It can be observed that our S$^4$TLG has a significant improvement compared with the baseline for both regression-based and proposal-based models, highlighting that our proposed components can effectively utilize the coarsely annotated data to improve grounding performance. 
Especially for the regression-based model, our S$^4$TLG boosts the baseline from 5.62 to 8.16 and from 6.26 to 9.15 in terms of dR@1,IoU@0.7 on ActivityNet-CD-OOD and Charades-CD-OOD, respectively.
Meanwhile, S$^4$TLG achieves comparable results to the baseline that uses all labeled data  (baseline@100\%) for both types of models. The results indicate that S$^4$TLG can effectively utilize unlabeled data for learning better features via contrastive losses and generating high-quality pseudo labels. In addition, on both datasets, the regression-based model achieves more significant improvements than proposal-based model. One possible reason is that the proposal-based model relies more on the quality of labels and predefined anchor scales, thus is more sensitive to the noise in pseudo labels.

\begin{table*}[t]
	\centering
	\footnotesize
	\begin{tabular}{cccccccccc}
		\toprule
		Row & Pre-train & Pseudo  & \begin{tabular}[c]{@{}c@{}}Sequential\\ Perturbations\end{tabular} & \begin{tabular}[c]{@{}c@{}}Intra-modal\\ Contrastive Loss\end{tabular} & \begin{tabular}[c]{@{}c@{}}Inter-modal\\ Contrastive Loss\end{tabular} & \begin{tabular}[c]{@{}c@{}}dR@1,\\ IoU@0.1\end{tabular} & \begin{tabular}[c]{@{}c@{}}dR@1,\\ IoU@0.3\end{tabular} & \begin{tabular}[c]{@{}c@{}}dR@1,\\ IoU@0.5\end{tabular} & \begin{tabular}[c]{@{}c@{}}dR@1,\\ IoU@0.7\end{tabular} \\
		\midrule
		1   & $\surd$   &         &                                &                                &                                & 43.14                          & 27.56                          & 15.41                          & 5.62                           \\
		2   & $\surd$   & $\surd$ &                                &                                &                                & 47.71                          & 29.41                          & 16.36                          & 6.80                           \\
		3   & $\surd$   & $\surd$ & $\surd$                        &                                &                                & 48.78                          & 30.05                          & 16.70                          & 6.94                           \\
		4   & $\surd$   & $\surd$ & $\surd$                        & $\surd$                        &                                & 48.14                          & 30.27                          & 17.14                          & 7.25                           \\
		5   & $\surd$   & $\surd$ &                                &                                & $\surd$                        & 48.88                          & 30.70                          & 17.47                          & 7.53                           \\
		6   & $\surd$   & $\surd$ & $\surd$                        &                                & $\surd$                        & \textbf{50.75}                 & 32.14                          & 17.86                          & 7.69                           \\
		7   & $\surd$   & $\surd$ & $\surd$                        & $\surd$                        & $\surd$                        & 50.71                          & \textbf{32.40}                 & \textbf{18.70}                 & \textbf{8.16}                  \\ \bottomrule
	\end{tabular}
	\caption{Ablation study of regression-based method on ActivityNet-CD-OOD.}
	\label{tab:label2}
	\vspace{-2mm}
\end{table*}
\begin{table*}[t]
	\centering
	\footnotesize
	\begin{tabular}{ccccccccc}
		\toprule
		Row & Pre-train & Pseudo  & \begin{tabular}[c]{@{}c@{}}Sequential\\ Perturbations\end{tabular} & \begin{tabular}[c]{@{}c@{}}Intra-modal\\ Contrastive Loss\end{tabular} & \begin{tabular}[c]{@{}c@{}}Inter-modal\\ Contrastive Loss\end{tabular} & \begin{tabular}[c]{@{}c@{}}dR@1,\\ IoU@0.3\end{tabular} & \begin{tabular}[c]{@{}c@{}}dR@1,\\ IoU@0.5\end{tabular} & \begin{tabular}[c]{@{}c@{}}dR@1,\\ IoU@0.7\end{tabular} \\
		\midrule
		1   & $\surd$   &         &                                &                                &                                & 28.91                          & 15.44                          & 6.16                           \\
		2   & $\surd$   & $\surd$ &                                &                                &                                & 28.84                          & 15.31                          & 6.39                           \\
		3   & $\surd$   & $\surd$ & $\surd$                        &                                &                                & 28.95                          & 16.10                          & 7.04                           \\
		4   & $\surd$   & $\surd$ & $\surd$                        & $\surd$                        &                                & 30.12                          & 16.64                          & 7.08                           \\
		5   & $\surd$   & $\surd$ &                                &                                & $\surd$                        & 30.48                          & 16.90                          & 7.30                           \\
		6   & $\surd$   & $\surd$ & $\surd$                        &                                & $\surd$                        & 30.63                          & 17.41                          & 7.84                           \\
		7   & $\surd$   & $\surd$ & $\surd$                        & $\surd$                        & $\surd$                        & \textbf{31.71}                 & \textbf{17.41}                 & \textbf{8.08}                  \\ \bottomrule
	\end{tabular}
	\caption{Ablation study of proposal-based method on ActivityNet-CD-OOD.}
	\label{tab:label3}
	\vspace{-4mm}
\end{table*}
\begin{table}[t]
	\centering
	\footnotesize
	\begin{tabular}{lccc}
		\toprule
		             & \begin{tabular}[c]{@{}c@{}}dR@1,\\ IoU@0.3\end{tabular} & \begin{tabular}[c]{@{}c@{}}dR@1,\\ IoU@0.5\end{tabular} & \begin{tabular}[c]{@{}c@{}}dR@1,\\ IoU@0.7\end{tabular} \\
		\midrule
		\multicolumn{4}{c}{regression-based}                                                                            \\
		\midrule
		 S$^4$TLG-comp & 31.77                          & 18.12                          & 8.02                           \\
		S$^4$TLG-lagging & 31.64                          & 18.20                          & 7.98                           \\
		S$^4$TLG         & \textbf{32.40}                          & \textbf{18.70}                          & \textbf{8.16}                           \\
		\hdrule
		\multicolumn{4}{c}{proposal-based}                                                                              \\
		\midrule
		S$^4$TLG-comp & 31.84                          & 17.45                          & 7.39                           \\
		S$^4$TLG-lagging & 29.40                          & 16.63                          & 7.24                           \\
		S$^4$TLG         & \textbf{31.71}                          & \textbf{18.00}                          & \textbf{8.08}                           \\ \bottomrule
	\end{tabular}
	\caption{Ablation study of perturbations on ActivityNet-CD-OOD. 
	}
	\label{tab:label4}
	\vspace{-7mm}
\end{table}
%-------------------------------------------------------------------------

\textbf{Comparisons with SOTA methods}. We then compare our S$^4$TLG with fully- and weakly-supervised state-of-the-art methods.
As Table~\ref{tab:label1} shows, while using limited labeled data, our S$^4$TLG significantly outperforms the weakly-supervised method WSSL~\cite{Duan2018} in all metrics.
Compared with fully-supervised methods, our regression-based model can achieve competitive performance on ActivityNet-CD-OOD and Charades-CD-OOD with only 5\% and 30\% data labeled with temporal boundary.
%-------------------------------------------------------------------------
\vspace{-3mm}
\subsection{Ablation Study}
\vspace{-2mm}
To show the effectiveness of each component,
we conduct ablation studies for both regression-based and proposal-based models on ActivityNet-CD-OOD dataset. The results are shown in Table~\ref{tab:label2} and Table~\ref{tab:label3}.

\textbf{Pseudo Label Generation.}
The performance of S$^4$TLG for regression-based model with and without pseudo label generation is shown in Table~\ref{tab:label2} (the 1$^{st}$ and 2$^{nd}$ row).
It can be observed that the pseudo label generation improves the temporal grounding accuracy by using instant pseudo labels. The improvements demonstrate the effectiveness of our pseudo label generating module. In Table~\ref{tab:label3} (the 1$^{st}$ and 2$^{nd}$ row) we also observe that using pseudo labels leads to performance improvement on dR@1,IoU@0.7, but the improvements are smaller than for regression-based models. We conjecture that the reason could be that proposal-based pseudo labels suffer from noise, so it is natural to introduce extra regularization to improve the teacher model and the label quality as well.

\textbf{Sequential Perturbations.}
Our designed perturbations, time lagging and video compression simulation, are motivated by the insight that moderate perturbation is beneficial for representation learning.
To investigate their effectiveness, we compare the S$^4$TLG's performance with and without sequential perturbations. The results are shown in Table~\ref{tab:label2} and Table~\ref{tab:label3} (the 2$^{nd}$ and 3$^{rd}$ row).
It is clear that the performance is improved by applying sequential perturbations to augment the training samples, and this validates that slightly altering the temporal structure of videos can help our model focus more on the semantics. 
We further investigate the effectiveness of individual sequential perturbation. As shown in Table~\ref{tab:label4}, both perturbations can improve the performance, and time lagging is slightly more effective than video compression simulation.

\textbf{Inter-modal Contrastive Loss.}
In order to validate the effectiveness of the inter-modal contrastive loss, we evaluate the S$^4$TLG with and without inter-modal contrastive loss.
By comparing the 2$^{nd}$ and 5$^{th}$ row in both Table~\ref{tab:label2} and Table~\ref{tab:label3}, we observe a significant performance improvement for all metrics, e.g., the dR@1,IoU@0.7 are improved by 10.7\% and 14\% for regression-based and proposal-based models, respectively. Same conclusion can be obatained by comparing the 3$^{rd}$ and 6$^{th}$ in Table~\ref{tab:label2} and Table~\ref{tab:label3}. The results suggest that the inter-modal contrastive loss can help to learn more discriminative visual and language features.
As we suggested earlier, inter-modal self-supervised learning can be used as an implicit regularization to improve the label quality especially for the proposal-based model, and the results here also agree with our previous conjecture.

\textbf{Intra-modal Contrastive Loss.}
We further investigate the effect of intra-modal contrastive loss. As shown in Table~\ref{tab:label2} and~\ref{tab:label3} (the 3$^{nd}$ and 4$^{th}$ row), when using intra-modal contrastive loss alone, the performance improves slightly. However, a substantial performance improvement can be obtained by combining intra-modal contrastive loss with inter-modal contrastive loss (the 6$^{th}$ and 7$^{th}$ row). The results suggest intra-modal contrastive loss plays a supplementary role to inter-modal contrastive loss in feature learning for temporal visual grounding task. Moreover, learning a better feature space that aligns visual and textual features well is the primary requirement for cross-modal tasks, and it can be the basis for further improving the feature learning within a particular modality.

%-------------------------------------------------------------------------
\vspace{-4mm}
\section{Conclusion}
\vspace{-2mm}
In this paper, we introduced a semi-supervised temporal language grounding framework assisted by contrastive self-supervised learning.
We designed pseudo label generation mechanisms that fit different types of grounding models and can be used in an instant updating manner during semi-supervised learning.
We also presented two effective sequential perturbations, and associated with contrastive self-supervised learning, which can further boost the performance of S$^4$TLG.
Our experiments show that S$^4$TLG achieves competitive performance compared with fully-supervised methods using a limited amount of temporal annotations, and S$^4$TLG is universally effective for different types of temporal language grounding models.
	%-------------------------------------------------------------------------

{\small
	\bibliographystyle{ieee_fullname}
	\bibliography{egbib}
}

\end{document}